\documentclass[twoside]{article}

\usepackage[accepted]{aistats2022}
\usepackage[round]{natbib}

\usepackage{graphicx}
\usepackage{algorithm}
\usepackage{algorithmic}
\usepackage{comment}
\usepackage{caption}
\usepackage{subcaption}
\usepackage{color}
\usepackage[implicit=false]{hyperref}
\usepackage{url}
\usepackage{amsthm}
\usepackage{amsfonts}
\newtheorem{theorem}{Theorem}
\newtheorem{corollary}{Corollary}[theorem]

\begin{document}

%
\runningtitle{Multiple Importance Sampling ELBO and
Deep Ensembles of Variational Approximations}

%
\runningauthor{Oskar Kviman, Harald Melin, Hazal Koptagel, Víctor Elvira, Jens Lagergren}

\twocolumn[

\aistatstitle{Multiple Importance Sampling ELBO and \\
Deep Ensembles of Variational Approximations}

\aistatsauthor{ Oskar Kviman$^{1,2}$ \And Harald Melin$^{1,2}$ \And  Hazal Koptagel$^{1,2}$ \And Víctor Elvira$^{3}$ \And Jens Lagergren$^{1,2}$}

\aistatsaddress{ \thanks{}KTH Royal Institute of Technology \And \thanks{}Science for Life Laboratory \And \thanks{}University of Edinburgh} ]

\newcommand{\cred}{\textcolor{red}}
\newcommand{\cblue}{\textcolor{blue}}

\begin{abstract}
In variational inference (VI), the marginal log-likelihood is estimated using the standard evidence lower bound (ELBO), or improved versions as the importance weighted ELBO (IWELBO). We propose the multiple importance sampling ELBO (MISELBO), a \textit{versatile} yet \textit{simple} framework. MISELBO is applicable in both amortized and classical VI, and it uses ensembles, e.g., deep ensembles, of independently inferred variational approximations. As far as we are aware, the concept of deep ensembles in amortized VI has not previously been established.
We prove that MISELBO provides a tighter bound than the average of standard ELBOs, and demonstrate empirically that it gives  tighter bounds than the average of IWELBOs. MISELBO is evaluated in density-estimation experiments that include MNIST and several real-data phylogenetic tree inference problems. First, on the MNIST dataset, MISELBO boosts the density-estimation performances of a state-of-the-art model, nouveau VAE. Second, in the phylogenetic tree inference setting, our framework enhances a state-of-the-art  VI algorithm that uses normalizing flows. On top of the technical benefits of MISELBO, it allows to unveil connections between VI and recent advances in the importance sampling literature, paving the way for further methodological advances. We provide our code at \url{https://github.com/Lagergren-Lab/MISELBO}.
\end{abstract}
\begin{figure}
    \centering
    \includegraphics[width=0.45\textwidth]{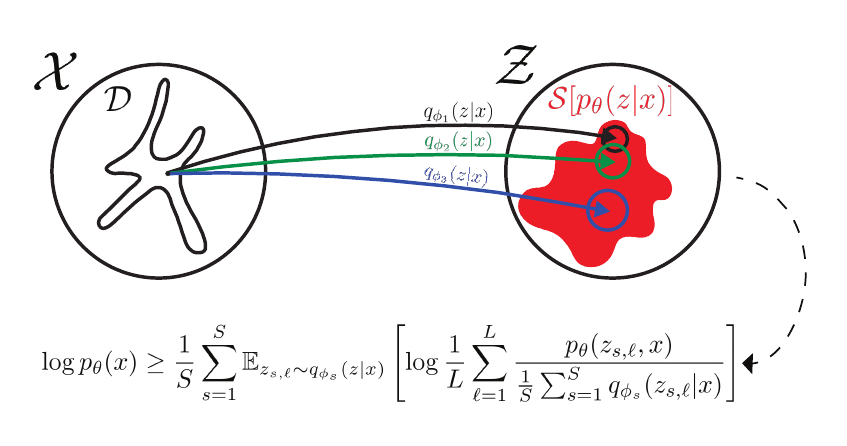}
    \caption{Our proposed framework. First, a set of variational approximations are independently obtained in the latent space above. Second, the marginal log-likelihood is estimated using MISELBO and the ensemble of variational approximations.}
    \label{fig:mappings}
\end{figure}
\section{Introduction}
\label{sec:intro}
Variational inference (VI;  \cite{jordan1999introduction,blei2017variational}) is an optimization-based approach to probability density estimation. An intractable posterior distribution, $p_\theta(z|x)$, is approximated by a variational approximation, $q_{\phi}(z|x)$, via maximizing of an objective function, typically the standard ELBO,
\begin{equation}
    \mathcal{L} = \mathbb{E}_{z\sim q_\phi(z|x)}\left[\log\frac{p_\theta(x,z)}{q_\phi(z|x)}\right],
\end{equation}
where $\phi$ and $\theta$ are the variational and generative model parameters, respectively.
In classical VI, $\phi$ is uniquely inferred for every data point $x$ in the dataset, $\mathcal{D}$, rendering it computationally expensive  \citep{zhang_vi_review}. In contrast, amortized VI is based on learning a \textit{mapping}, $f _\phi(x)$, from the data $\mathcal{D}$ to the parameters of the approximate posterior, 
 which is applied across all data points. This makes amortized VI efficient and preferable for large-scale problems. Typically, $f_\phi(x)$ is a neural network (NN), and its weights, $\phi$, are
 learned via stochastic gradient descent (SGD) updates.

The variational auto-encoder (VAE;  \citet{kingma2013auto,rezende2014stochastic}) is an important class of amortized VI algorithms.
During training, the variational and model parameters, $\phi$ and $\theta$, are jointly learned via SGD optimization of the objective function. 

Recently, there has been a surge of research regarding alternative objective functions  \citep{higgins2016beta, kim2018disentangling, sinha2021consistency}. Some of these results are based on divergence measures  \citep{li2016r, dieng2017variational,wang2018variational, tran2021cauchy}, and while others have proposed tighter lower bounds \citep{masrani2019thermodynamic}. Especially,  \cite{burda2015importance} proposed the importance weighted ELBO (IWELBO)
\begin{equation}
\label{eq:iwelbo}
    \mathcal{L}_L= \mathbb{E}_{z_{1},...,z_{L}\sim q_{\phi}(z|x)}\left[
    \log\frac{1}{L}\sum_{\ell=1}^L \frac{p_{\theta}(z_{\ell},x)}{ q_{\phi}(z_{\ell}|x)}
    \right],
\end{equation}
which has been extensively used as an objective function  \citep{burda2015importance, sonderby2016ladder, aitchison2018tensor, lopez2020decision} and, importantly,   as a metric for estimating the marginal log-likelihood, $\log p_\theta(x)$, in VAEs (e.g.,  \cite{tomczak2018vae, bauer2019resampled, vahdat2020nvae}) and in VI in general  \citep{domke2018importance, zhang2020improved}. 

There are some limiting properties imposed upon $q_\phi(z|x)$: it should be easy to sample from (preferably through reparameterization), and its likelihood must be tractable. These properties often constrain $q_\phi(z|x)$ to a unimodal family of distributions, making it a simplistic approximation of the intractable posterior. While there have been successful attempts to learn multimodal and other expressive variational approximations, e.g., via the introduction of auxiliary variables in  \cite{maaloe2016auxiliary} or normalizing flows  (NF), as in  \cite{rezende2015variational,kingma2016improved}, these approaches can usually not be straightforwardly be applied to existing methods or require practitioners to expertise in certain methodologies. 

In this paper, we propose a flexible framework for obtaining an ensemble of independently inferred variational approximations, enabling the practitioner to employ their preferred VI algorithm and still obtain a multimodal variational approximation. 
We derive the alternative multiple importance sampling ELBO (MISELBO), which uses the ensemble, as
\begin{align}
\label{eq:miselbo}
    \mathcal{L}^L_\text{MIS}\mathcal =\frac{1}{S}\sum_{s=1}^S \mathbb{E}_{ q_{\phi_s}(z|x)}\left[
    \log\frac{1}{L}\sum_{\ell=1}^L \frac{p_{\theta}(z_{s,\ell},x)}{ \frac{1}{S}\sum_{s=1}^Sq_{\phi_s}(z_{s,\ell}|x)}
    \right],
\end{align}
where $z_{s,\ell}\sim q_{\phi_s}(z|x)$. Our framework is visualized in Figure \ref{fig:mappings}.

The MISELBO is motivated by recent advances in the importance sampling (IS) community, and more particularly in the context of multiple IS (MIS), where multiple proposals/approximations are available, as it is the case here. Recently, in  \cite{elvira2019generalized}, it was  shown that, using the mixture weights in Eq.  \eqref{eq:miselbo} (also called balance heuristic  \citep{veach1995} or deterministic mixture  \citep{owen2000}) always provides better estimators in terms of variance than using the standard weights in Eq. \eqref{eq:iwelbo}. In the VI framework, where the ELBO computes an expectation of the log transformation, the improvement of the mixture weights is translated into a tighter bound than when the standard ones are used. These connections have promising implications for future research.


Our framework can be easily applied in the context of deep learning since, as in importance sampling (IS) (see \citep{elvira2021advances}), the set of $q_{\phi_s}(z|x)$, proposal distributions,  can be independently inferred/trained despite sharing the same $p_\theta(z, x)$, target distribution. Consequently, we can exploit deep ensemble diversity  \citep{lakshminarayanan2016simple, fort2019deep}, which  is a novel insight, as far we are aware. 

In short, our contributions are:
\begin{itemize}
    \item We prove that MISELBO is tighter than the average of standard ELBOs, $\overline{\mathcal{L}}$, and show through experiments that it is also tighter than the IWELBO (Section \ref{sec:tightness miselbo}).
    \item We establish the concept of deep ensembles of variational approximations (Section \ref{deep ensembles in avi}).
    \item We propose a framework which utilizes ensembles of variational approximations in order to improve marginal log-likelihood estimates for existing algorithms, e.g., NVAE  (\cite{vahdat2020nvae}; Section \ref{sec:miselbo} and \ref{sec:experiments}).
    \item We show that the multimodal posterior distribution in the NN weight space  \citep{wilson2020bayesian}, known to be induced by deep ensemble diversity  \citep{fort2019deep}, translates into a multimodal ensemble of variational approximations in the latent space (Section \ref{sec:mnist}).
\end{itemize}

\section{Background}
\label{sec:background}

Deep ensembles  \citep{lakshminarayanan2016simple}, are known to boost prediction performance and uncertainty quantification, while being easy to train. The practitioner simply trains $S$ models independently in parallel, with randomly initialized network weights. Indeed, random initialization provides many opportunities.  For instance, \cite{fort2019deep} found that it forces the deep NNs to explore different modes in the NN weight space, making deep ensembles \textit{diverse}. Outside the neural network setting, \cite{YaoStacking} showed that diverse variational approximations may be learned in the classical mean-field setting. This was achieved via initialization of the approximations in different parts of the parameter space, and optimizing them using stochastic gradient ascent. In this work, we leverage these findings to obtain diverse ensembles of independently trained variational approximations to boost performance.

In information theory, the Jensen-Shannon divergence (JSD) is a non-negative, symmetric divergence measure for a set of probability distributions, $\mathcal{Q_S}=\{q_{\phi_s}(z|x)\}_{s=1}^S$. Assuming the distributions are equally weighted, the JSD is formulated as
\begin{equation}
\label{eq:jsd}
    \text{JSD}(\mathcal{Q}_S) = \mathbb{H}\left[
    \frac{1}{S}\sum_{s=1}^S q_{\phi_s}(z|x)
    \right] - \frac{1}{S}\sum_{s=1}^S\mathbb{H}\left[
    q_{\phi_s}(z|x)
    \right],
\end{equation}
where $\mathbb{H}[\cdot]$ is the entropy function. Furthermore,  the JSD is upper- and lower-bounded. In fact $\text{JSD}(\mathcal{Q}_S)\in [0, \log S]$. As we show in Section \ref{sec:experiments},  the JSD is suitable for measuring the diversity of an ensemble of variational approximations. The benefits of obtaining diverse ensembles in Eq. \eqref{eq:miselbo} is motivated by recent work in the MIS literature. Namely,  \cite{elvira2019generalized} showed that the improvement of the MIS weights are more effective (versus the standard weights) when each approximation
$q_{\phi_s}(z|x)$ 
is significantly different with respect to the mixture (high JSD), which aims at mimicking the target distribution. Meanwhile the MIS weights do not provide any improvement when all $q_{\phi_s}(z|x)$  are identical (JSD $=0$). This motivates using the JSD to measure the effectiveness of the ensemble.

We use the \textit{average of ELBOs}
\begin{equation}
\label{eq:avgELBOs}
    \overline{\mathcal{L}} = \frac{1}{S}\sum_{s=1}^S \mathbb{E}_{q_{\phi_s}(z|x)}\left[
    \log \frac{p_{\theta}(z_{s},x)}{ q_{\phi_s}(z_{s}|x)}
    \right],
\end{equation}
and the \textit{average of IWELBOs}
\begin{equation}
    \overline{\mathcal{L}}_L = \frac{1}{S}\sum_{s=1}^S \mathbb{E}_{q_{\phi_s}(z|x)}\left[
    \log\frac{1}{L}\sum_{\ell=1}^L \frac{p_{\theta}(z_{s,\ell},x)}{ q_{\phi_s}(z_{s,\ell}|x)}
    \right],
\end{equation}
as proxies for the ELBO and IWELBO, respectively. This observation makes it simpler for us to  compare MISELBO with the other two lower bounds. Indeed, in the setting of deep ensembles, we found that for a set of IWELBOs the standard deviation  is small, especially for big $L$ (e.g., see Figure \ref{fig:NLL_v_L_mnist}).

\section{Multiple Importance Sampling Evidence Lower Bound}
\label{sec:miselbo}

\begin{algorithm}[tb]
\small
   \caption{Pseudocode for estimating the MISELBO in Eq. \eqref{eq:miselbo}}
   \label{alg:miselbo}
\begin{algorithmic}
    \STATE {\bfseries Inputs:} $p_\theta(z, x),q_{\phi_1}(z|x),...,q_{\phi_S}(z|x)$
    \STATE Initialize $\tilde{\mathcal{L}} = 0$
    \FOR{$s=1,...,S$}
       \STATE $\{z_{s,l}\}_{l=1}^L\sim q_{\phi_s}(z|x)$
            \STATE $\tilde{\mathcal{L}}\gets \tilde{\mathcal{L}} + \frac{1}{S}\log\frac{1}{L}\sum_{\ell=1}^L \frac{p_{\theta}(z_{s,\ell},x)}{\frac{1}{S}\sum_{j=1}^S  q_{\phi_{j}}(z_{s,\ell}|x)}$
       
    \ENDFOR
\end{algorithmic}
\end{algorithm}

Let us, using $\mathcal{S}$ to denote the support of a distribution,  define a set of of variational approximations such that
\begin{equation}
\label{eq:set}
    \mathcal{Q}_S = \{
    q_{\phi_s}(z|x) :  \mathcal{S}\left[q_{\phi_s}(z|x)\right] \subseteq \mathcal{S}\left[p_\theta(z|x)\right]
    \}_{s=1}^S.
\end{equation}
Importantly, note that the constraint in Eq. \eqref{eq:set} occurs naturally when obtaining $q_{\phi_s}(z|x)$ via minimization of $\text{KL}(q_{\phi_s}(z|x)\Vert p_\theta(z|x))$. The constraint means that all approximations must be absolutely continuous with respect to \textit{the same} target distribution, $p_\theta(z|x)$. For example in a VAE setting, the final training iteration for all encoder networks must be obtained using the same decoder network. Throughout the paper, all comparisons between $\overline{\mathcal{L}}_L$ and $\mathcal{L}_\text{MIS}^L$ are done using the same set $\mathcal{Q}_S$, unless mentioned otherwise.

When using our framework, the practitioner simulates a set of $S \times L$ samples (as in the case of the average IWELBOs), and then produces the estimator of the marginal log-likelihood using MISELBO in Eq. \eqref{eq:miselbo}.
Although we in this work sample $L$ times from each $q_{\phi_s}(z|x)$, and weight the samples and approximations equally, there are many possible ways of combining sampling and weighting schemes in the MIS literature  \citep{elvira2019generalized,sbert2022generalizing}. Sophisticated choices of these schemes can reduce the variance of the estimator, make the ensemble focus on high-probability regions, sample more economically, and more. Also, efficient schemes to find the optimal tradeoff between performance and complexity can be explored (e.g., in the lines of \cite{elvira2015efficient}). Hence, MISELBO paves the way for numerous interesting research directions by bridging the gap between VI and MIS.

\subsection{Tightness of MISELBO}
\label{sec:tightness miselbo}
In the following, we prove that MISELBO provides a tighter bound than the average  $\overline{\mathcal{L}}(\mathcal{Q}_S)$. First, let 
\begin{equation}
    \Delta_L = \mathcal{L}_\text{MIS}^L(\mathcal{Q}_S) -\overline{\mathcal{L}}_L(\mathcal{Q}_S)
\end{equation}
denote the difference between MISELBO and the average of IWELBOs, which with $L=1$ turns into the average of ELBOs and is   denoted $ \Delta_1$. 
\begin{theorem}
\label{thm:miselbo_v_elbo}
For a given set $\mathcal{Q}_S$, the following inequality holds
\begin{align*}
     \mathcal{L}_\text{MIS}(\mathcal{Q}_S) \geq \overline{\mathcal{L}}(\mathcal{Q}_S),
\end{align*}
and, since $\overline{\mathcal{L}}_1 \equiv \overline{\mathcal{L}}$, the difference, $\Delta_{1} = \mathcal{L}_\text{MIS}(\mathcal{Q}_S) - \overline{\mathcal{L}}(\mathcal{Q}_S) $, satisfies both the upper and lower bounds,
\begin{align*}
     \log S \geq\Delta_{1} \geq 0.
\end{align*}
\end{theorem}

\begin{proof}
We evaluate the difference directly
\begin{align*}
    \Delta_{1}&= 
    \frac{1}{S}\sum_{s=1}^S\mathbb{E}_{q_{\phi_s(z|x)}}\left[
    \log \frac{p_\theta(x, z)}{\frac{1}{S}\sum_{s=1}^S q_{\phi_s}(z|x)}\right]
    \\ 
    &\quad-\frac{1}{S}\sum_{s=1}^S\mathbb{E}_{q_{\phi_s(z|x)}}\left[\log \frac{p_\theta(x, z)}{q_{\phi_s}(z|x)}
    \right]\\
    &= -\frac{1}{S}\sum_{s=1}^S\mathbb{E}_{q_{\phi_s(z|x)}}\left[
    \log {\frac{1}{S}\sum_{s=1}^S q_{\phi_s}(z|x)}
    \right]\\ 
    &\quad+ \frac{1}{S}\sum_{s=1}^S\mathbb{E}_{q_{\phi_s(z|x)}}\left[
    \log {q_{\phi_s}(z|x)}
    \right]\\
    &= \mathbb{H}\left[
    \frac{1}{S}\sum_{s=1}^S q_{\phi_s}(z|x)
    \right] - \frac{1}{S}\sum_{s=1}^S\mathbb{H}\left[
    q_{\phi_s}(z|x)
    \right]\\ &= \text{JSD}(\mathcal{Q}_S) \geq 0.
\end{align*}
From Section \ref{sec:background}, we know that JSD$(\mathcal{Q}_S)\in[0, \log S]$, and so the proof is complete.
\end{proof}
    
\begin{corollary}
\label{corollary strict inequality}
The inequality is strict when $\bigcap_{s=1}^S \mathcal{S}\left[q_{\phi_s}(z|x)\right] = \emptyset$, 
$\mathcal{L}_\text{MIS}(\mathcal{Q}_S) > \overline{\mathcal{L}}(\mathcal{Q}_S)$.
\end{corollary}
\begin{proof}
 See supplementary material.
\end{proof}

\begin{corollary}
\label{corollary equality}
There is equality when all distributions in $\mathcal{Q}_S$ are the same, 
$\mathcal{L}_\text{MIS}(\mathcal{Q}_S) = \overline{\mathcal{L}}(\mathcal{Q}_S)$.
\end{corollary}
\begin{proof}
See supplementary material.
\end{proof}

Moreover, as \cite{burda2015importance} does not restrict the variational approximation to any certain family of distributions in their results, the following also holds for MISELBO (where the variational approximation is an equally weighted ensemble)
\begin{align}
    \mathcal{L}^L_\text{MIS} \geq \mathcal{L}^{L-1}_\text{MIS} \geq  \mathcal{L}^1_\text{MIS},
\end{align}
and 
\begin{equation}
    \log p_\theta(x) \geq \mathcal{L}^L_\text{MIS}.
\end{equation}

Although we have not proven that $\Delta_L$ is strictly non-negative, this is what our experiments, presented in Section \ref{sec:experiments},  consistently indicate.

\subsection{Deep Ensembles in Amortized Variational Inference}
\label{deep ensembles in avi}

In the previous section, we showed that $\Delta_{1}$ increases with the JSD$(\mathcal{Q}_S)$. Although diversity is a vague concept, we hypothesize that the diversity of an ensemble of mappings $\{f_{\phi_s}(x)\}_{s=1}^S$ can be measured in the latent space using JSD$(\mathcal{Q}_S)$. Consequently, $\Delta_{1}$ grows as the ensemble of variational approximations becomes more diverse. Reversely, if we obtain a larger JSD$(\mathcal{Q}_S)$ simply by using a deep ensemble of independently trained variational approximations, then diversity in deep ensembles would promote multimodal posterior distributions, not only over the NN weights \cite{wilson2020bayesian}, but also in the latent space. Measuring deep ensemble diversity in the latent space, appears to be a novel idea.

As mentioned in Section \ref{sec:intro}, VAEs are important instances of amortized VI, and so we apply them to deep ensembles of variational approximations in Section \ref{sec:mnist}. In order for deep ensembles of variational approximations to be applicable, however, there needs to be a single mapping from the latent space to the likelihood parameters, $f_\theta(z)$. We solve this by, first, training a VAE, obtaining $\{f_{\phi_1}(x), f_{\theta}(z)\}$, and then fixing $f_{\theta}(z)$ --- no gradient updates with respect to $\theta$ are computed at this point. We now have a generative model, $p_\theta(x, z)$, to use to independently train the $S-1$ other $f_{\phi_s}(x)$, or, equivalently, for inferring $\{q_{\phi_s}(z|x)\}_{s=2}^S$. Algorithm \ref{alg:misvae} displays the simplicity of the framework.

We refer to a deep ensemble of independently trained mappings $\{f_{\phi_s}(x)\}_{s=1}^S$ as a deep ensemble of variational approximations, $\mathcal{Q}_S$, and their mappings are learned using the same decoder, satisfying $\mathcal{S}[q_{\phi_s}(z|x)]\subseteq \mathcal{S}[p_{\theta}(z|x)]$ for all $q_{\phi_s}(z|x)\in\mathcal{Q}_S$. 

\begin{algorithm}[tb]
\small
   \caption{Pseudocode for deep ensembles of variational approximations. Here, $\mathcal{H}(\cdot, \cdot)$ is the practitioners choice of objective function}
   \label{alg:misvae}
\begin{algorithmic}
\STATE {\bfseries Inputs:} $f_{\phi_1}(x), f_\theta(z), \mathcal{D}$
    \STATE  Initialize  $\phi_2,...,\phi_S$ randomly
    \FOR{$s=2,...,S$}
       \STATE Train $f_{\phi_s}(x)$ on $\mathcal{D}$ via $\partial_{\phi_s} \mathcal{H}\left(f_{\phi_s}(x), f_\theta(z)\right)$ in parallel
    \ENDFOR 
\end{algorithmic}
\end{algorithm}

\section{Related Work}
\label{sec:variational approximations}
\begin{figure*}
    \centering
    \begin{subfigure}[b]{0.35\textwidth}
    \centering
    \includegraphics[width=0.85\textwidth]{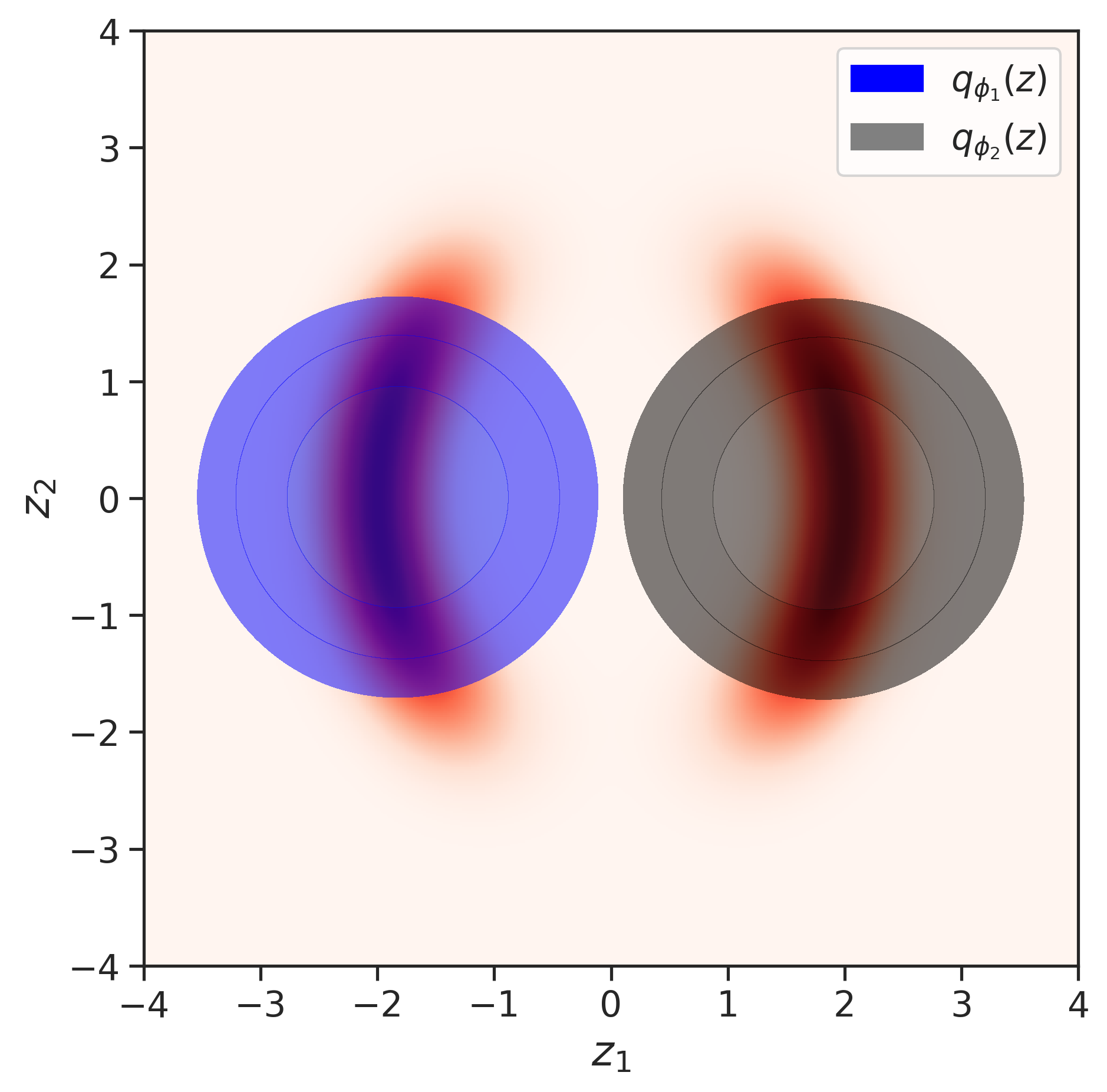}
    \caption{$p_1(z)$}
    \end{subfigure}\hspace*{1cm}%
    \begin{subfigure}[b]{0.35\textwidth}
    \centering
    \includegraphics[width=0.85\textwidth, height=5cm]{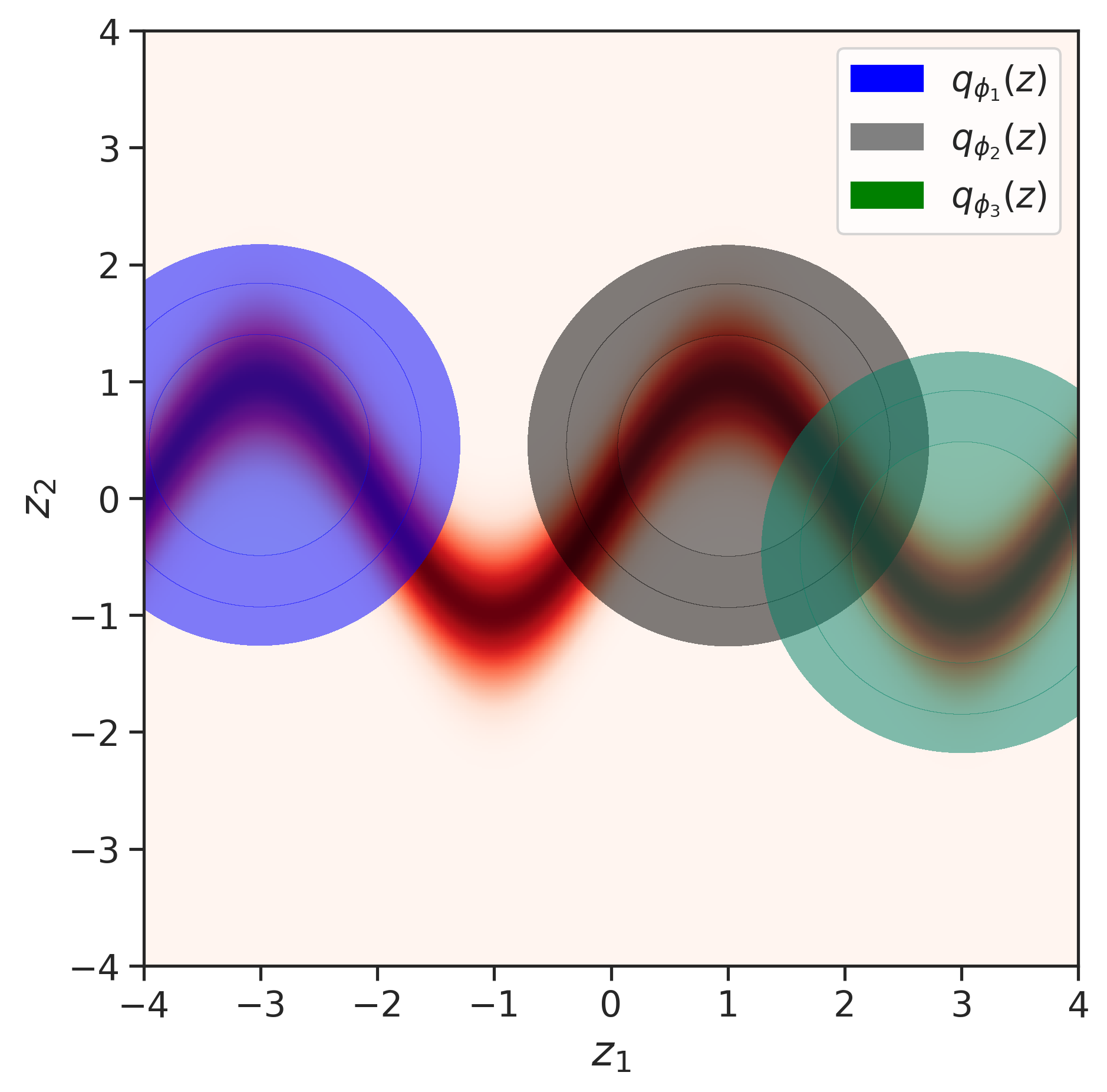}
    \caption{$p_2(z)$}
    \end{subfigure}
    \caption{The two true distributions,   in (a) and (b) are represented as heat maps. The variational approximations are the contour distributions. }
    \label{fig:vi_ens_approx}
\end{figure*}
\label{sec:related-work}
In recent years, some examples of ensembles of variational approximations and multimodal variational approximations have been proposed. The most important work relating to ours is that of \cite{lopez2020decision}, as their framework can result in an ensemble of encoder networks with a single decoder. The ensemble is then used for density estimation in interesting decision-making settings. To obtain their variational approximations and the decoder, they follow a three-step procedure which, briefly summarized, involves using different objective functions and selecting which $p_\theta(x,z)$ to use among a set of learned generative models. This is a different approach to ours, and their is not easily applied to the experiments conducted here. Our framework, deep ensembles of variational approximations, is flexible, principled and a generalization of their framework. Finally, they do not show how to use the ensemble to compute the evidence lower bound. We do this using the MISELBO. 

In \cite{hernandez2016importance} and \cite{daxberger2019bayesian} they work with ensembles of variational autoencoders, viewed as a Bayesian deep learning setup by obtaining distributions over the network parameters. They either either (i) obtain both $q(\phi)$ \textit{and} $q(\theta)$, or (ii) learn a single encoder (point mass $\phi$) and  $q(\theta)$. Clearly, (i) is the scheme most similar to our framework, albeit substantially different. They learn pairs of jointly trained encoders and decoders, making $q_{\phi_j}(z|x)$, for any $j\neq s$, not necessarily absolutely continuous with respect to $p_{\theta_s}(z|x)$. This violates the condition from Sec. \ref{sec:miselbo}, leaving MISELBO undefined in (i) and (ii).

Recently, \cite{thin2021monte} combine annealed importance sampling and sequential Monte Carlo methods to obtain better estimates of the marginal likelihood in the VAE setting. This is indeed an interesting direction which relates to our work by also being on the front line of importance sampling research. Since it is not clear how to apply this method to hierarchical VAEs or outside the VAE setting, and the resulting estimate of the marginal log-likelihood is only proven to empirically be tighter than the IWELBO, we do not compare with it here.

 \cite{guo2016boosting} introduced boosting VI, where ensemble components from a simple parametric base model are iteratively combined to create a more complex posterior. Also, discrete particle variational inference  \citep{saeedi2017variational} utilizes an ensemble of variational approximations. However, in both methods the components are jointly trained, and neither of them offer obvious extensions to the deep learning setting.  

For deep latent variable models, the prior of the generative model has been the predominant target of multimodal endeavours \citep{bauer2019resampled,bozkurt2021rate, tran2021cauchy}.
For VAEs, \cite{tomczak2018vae} introduced a mixture distribution prior by leveraging an aggregate posterior over pseudo-inputs, and \cite{jiang2016variational} a GMM prior by adding a categorical latent variable. Since these approaches relate to the prior distribution, they are not competing approaches, albeit well-worth mentioning due to their multimodal latent space assumptions.

A GMM variational posterior is provided in the deep latent Gaussian Mixture model of \cite{nalisnick2016approximate} and in the Varitiational Information Bottleneck of \cite{uugur2020variational}; both of these approaches rely on jointly optimized components of the GMMs.

\begin{table}
\begin{center}
\caption{The two multimodal distributions used in Section \ref{sec:variational approximations}.}\label{table:multimodal distributions}
\begin{tabular}{l}
\hline
$\log p_1(z) \propto -\frac{1}{2}\left( \frac{\Vert z\Vert - 2}{0.4}\right)^2 - \left(e^{-\left(
\frac{z_1 - 2}{1.2}
\right)} + e^{-\left(
\frac{z_1 +2}{1.2}
\right)}\right)$ \\
$\log p_2(z)\propto - \frac{1}{2} \frac{(z_2 - w(z))^2}{0.4}$, with $w(z) = \sin\left( \frac{2\pi z_1}{4} \right)$\\
\end{tabular}
\end{center}
\end{table}
\begin{table}
\centering
\caption{Unnormalized KL divergences and JSD.}\label{table:kl scores}
\begin{tabular}{cccc}
 &$\text{KL}_\text{MIS}$ & $\overline{\text{KL}}$ & \text{JSD}  \\\hline
$p_1(z)$& -0.03 & 0.61 & 0.64 \\  
$p_2(z)$& 0.15 & 1.05  & 0.90
\end{tabular}
\end{table}

\begin{figure*}
    \centering
    \begin{subfigure}[b]{1\textwidth}
    \centering
    \includegraphics[width=1\textwidth]{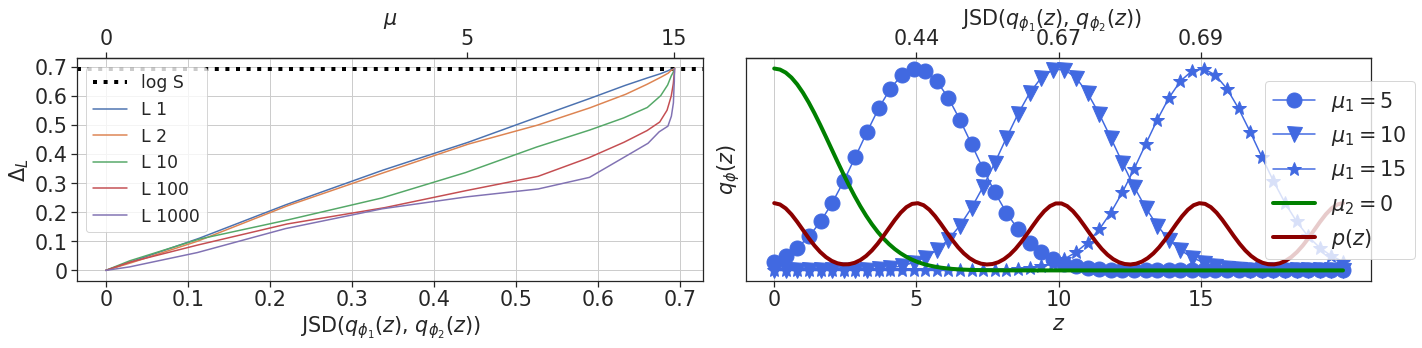}
    \end{subfigure}
    \caption{\textbf{Left}: Visualization of how $\Delta_L$ increases with the JSD$(\mathcal{Q}_S)$. Note that, regardless of $L$, $\Delta_L$ converges to $\log S$ ($\log2\approx 0.69$) as the two Gaussians are separated, or, equivalently, the JSD$(\mathcal{Q}_S)$ grows. Three examples of $\mu$ for $q_{\phi_1}(z)$ are shown in the top labels. \textbf{Right}: Examples of how shifting $q_{\phi_1}(z)$ (blue with $\mu_1=5,10,15$) away from $q_{\phi_2}(z)$ (the green with $\mu_2=0$) affects the JSD$(\mathcal{Q}_S$). The three corresponding JSD values are shown in the top labels.}
    \label{fig:jsd_v_mu}
\end{figure*}

In \cite{shi2019variational} they propose mixture of experts multimodal VAE (MMVAE), which spans multiple \textit{data modalities}, such as vision and language. Namely, the latent space is decomposed into multiple modes, each representing the corresponding data type. Conversely, the multimodality in our framework is w.r.t. the latent space as one component. Finally, MMVAE applies stratified sampling instead of important sampling.

The insights regarding deep ensembles in \cite{lakshminarayanan2016simple} are essential for out amortized VI variant. The NNs are independently trained, but concern discriminative networks.

\section{Experiments}
\label{sec:experiments}
We consider four density-estimation tasks. The first two experiments concern one- or two-dimensional distributions, in order to easier visualize our framework's strengths. Two large-scale experiments are then considered, displaying how MISELBO enables the use of deep ensembles in amortized VI, as well as its applicability to modern VI techniques such as VAEs and NF.

\subsection{Representative Power of the Proposed Framework}
To more easily display the power of our proposed method, we here consider two low-dimensional density estimation tasks.
\subsubsection{Ensembling Variational Approximations}
\label{subsec:ens_var_app}
We use two of the unnormalized and multimodal/periodic distributions described in  \citet{rezende2015variational} and defined in  Table \ref{table:multimodal distributions}. These distributions are known to be too complicated to fit using any standard approximate distribution. This experiment allows us to show (i) that, inspired by \cite{YaoStacking}, variational approximations with different initializations may ultimately cover different parts of $p(z)$, and (ii) how the resulting ensemble diversity can be  leveraged. 

\begin{figure}
    \centering
    \includegraphics[width=0.45\textwidth]{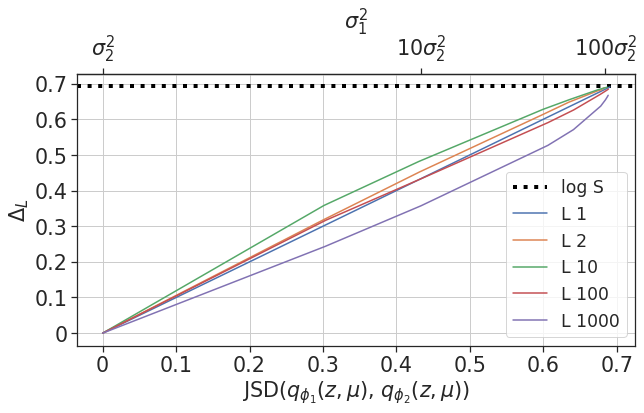}
    \caption{Visualization of how $\Delta_L$ increases with the JSD$(\mathcal{Q}_S)$ in a hierarchical model example (see Section \ref{sec:experiments:jsd} for details). Here we make the ensemble of variational approximations more diverse by increasing the variance in one of approximations, $\sigma_1^2$. Three examples of $\sigma_1^2$ for $q_{\phi_2}(\mu)$ are shown in the top labels.}
    \label{fig:jsd_v_std}
\end{figure}

\begin{table*}[!ht]
    \centering
    \begin{tabular}{l|c|c|c|c|c||c}
         & $L=1$ & $L=2$ & $L=50$ & $L=500$ & $L=1000$ & JSD($\mathcal{Q}_S)$\\\hline
         $\mathcal{L}^L_\text{MIS}$& $\bold{79.10} \pm 0.2$  & $\bold{79.00} \pm 0.2$ & \color{red} $\mathbf{78.12} \pm 0.2$ & $\bold{77.81} \pm 0.2$ & $\bold{77.77} \pm 0.2$  \\
         $\overline{\mathcal{L}_L}$ & $79.54 \pm 0.2$ & $79.43 \pm 0.2$ & $78.56 \pm 0.1$ & $78.25 \pm 0.1$ & \color{red} $78.21 \pm 0.1$  \\
         $\Delta_L$ & $0.42 \pm 0.24$ & $0.43 \pm 0.25$ & $0.44 \pm 0.25$ & $0.44 \pm 0.25$ & $0.44 \pm 0.25$ & $0.44 \pm 0.25$ \\
         Best $\mathcal{L}_L$ & $79.86 \pm 0.2$ & $79.29 \pm 0.2$ & $78.34 \pm 0.1$ & $78.20 \pm 0.1$ & \color{red} $78.19 \pm 0.1$ 
    \end{tabular}
    \caption{NLL scores and $\Delta_L$ for NVAEs on the MNIST dataset when $S=2$. The results were averaged using five different random seeds. In this experiments we observed that $\Delta_L\geq$ JSD($\mathcal{Q}_S)$ (JSD($\mathcal{Q}_S)$ is constant in $L$). Furthermore, comparing the \textcolor{red}{red} entries, we note that the deep ensemble of NVAEs+MISELBO requires 90\% less importance samples to outperform NVAE using IWELBOs.}
    \label{table:S=2 MNIST}
\end{table*}

The obtained fit  is quantified in terms of the KL divergence.  Specifically, we compare the following two quantities
\begin{equation}
    \text{KL}_\text{MIS} = \text{KL}\left(\frac{1}{S}\sum_{s=1}^Sq_{\phi_s}(z)\Big\Vert p(z)\right),
\end{equation}
\begin{equation}
    \overline{\text{KL}} = \frac{1}{S}\sum_{s=1}^S\text{KL}(q_{\phi_s}(z)\Vert p(z)).
\end{equation}
We let $q_{\phi_s}(z)=\mathcal{N}(\mu_s, 0.8)$, and minimize the  KL$(q_{\phi_s}(z)\Vert p(z))$ with respect to the variational parameter $\mu_s$ using the Adam optimizer   \citep{kingma2014adam}.  The variational approximations are obtained independently, and the parameters are initialized in separate parts of $z$-space. The resulting scores are presented in Table \ref{table:kl scores} where we observe high JSDs (close to $\log S$) in both settings. This indicates that diverse ensembles were obtained via different parameter initializations. Also, we see a clear improvement when using the diverse ensembles of variational approximations over averaging the $S$ solutions.

\subsubsection{JSD as a Diversity Metric and Visualizing $\Delta_L$}
\label{sec:experiments:jsd}
Here we visualize, using an ensemble of two variational approximations, how $\Delta_L$ is affected by the JSD$(\mathcal{Q}_S$). We do this for two cases. First, we let the variational approximations be unimodal Gaussians with variable means, i.e. $\mathcal{Q}_S = \{q_{\phi_1}(z), q_{\phi_2}(z)\}$, where $q_{\phi_s}(z) = \mathcal{N}(z|\mu_s, 1)$. Second, we consider a hierarchical model by introducing a prior on the mean parameter, and let $\mathcal{Q}_S = \{q_{\phi_1}(z|\mu)q_{\phi_1}(\mu), q_{\phi_2}(z|\mu)q_{\phi_2}(\mu)\}$. In the latter setting, $q_{\phi_s}(\mu)=\mathcal{N}(\mu|10, \sigma_s^2)$ has variable variance. 

In both cases, we quantify the diversity of the two approximations using the JSD$(\mathcal{Q}_S$). In the first case, we control the diversity by shifting $q_{\phi_1}(z)$ away from $q_{\phi_2}(z)$. In the second case, we gradually increase the variance of $q_{\phi_1}(\mu)$, starting from $\sigma_1^2=\sigma_2^2$. By doing this, we indirectly parameterize $\Delta_L$ by the ensemble diversities.

In Figure \ref{fig:jsd_v_mu}, we present the results of the first case, where we let the true model, $p(z)$, have six modes. In the supplementary material we include the results of the same experiment with less number of modes for $p(z)$ (the results are similar). In the right plot of Figure \ref{fig:jsd_v_mu}, it can be observed how shifting the two variational approximations apart increases the JSD$(\mathcal{Q}_S$), and hence the ensemble diversity. Recall from Equation \eqref{eq:jsd} that JSD$(\mathcal{Q}_S$) is independent of both $L$ and the true model. The results corresponding to the second case are displayed in Figure \ref{fig:jsd_v_std}.

For both cases we confirm that, firstly, $\Delta_{1}$ follows Theorem \ref{thm:miselbo_v_elbo}, i.e. $\Delta_{1} = \text{JSD} (\mathcal{Q}_S)$, and, secondly, Corollaries \ref{corollary strict inequality} and \ref{corollary equality} hold in practice. Especially, we take note that the performance gain of using MISELBO over the averaged lower bounds increases as the ensemble becomes more diverse, irrespective of $L$. 

Most importantly, in all of the experiments $\Delta_L$ is strictly non-negative when JSD$(\mathcal{Q}_S) > 0$. This inequality is recurring in all of our experiments in this paper, and indicates that the inequality might be true in general. However, it remains to be proven.  

\subsection{MNIST}
\label{sec:mnist}
\begin{table}
\caption{Negative log-likelihood (NLL) results on the MNIST dataset when $L=1000$ and $S=2$. We compare the marginal log-likelihood estimates from NVAE when using our MISELBO and when using the IWELBO.}
\label{table:NLL-benchmarking}
\begin{center}
\begin{tabular}{ll}
\multicolumn{1}{c}{\bf Model w. lower bound}  &\multicolumn{1}{c}{\bf NLL}
\\ \hline \\
NVAE w. IWELBO      & $78.21\pm0.1$ \\
NVAE w. MISELBO             & $\mathbf{77.77\pm 0.2}$\\
\end{tabular}
\end{center}
\end{table}

\begin{figure}
    \centering
    \includegraphics[width=0.45\textwidth]{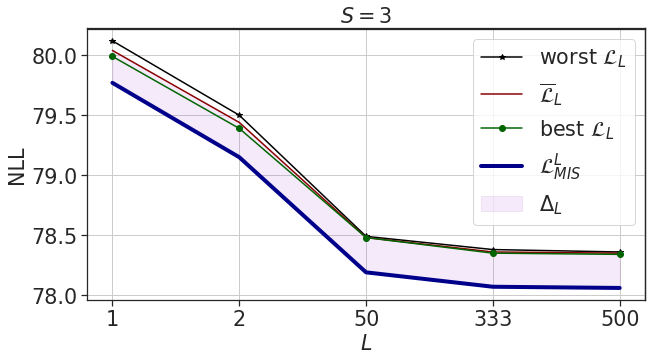}
    \caption{NLL curves and $\Delta_L$ for NVAEs and a single random seed on the MNIST dataset when $S=3$ and varying $L$. As $L$ increases the IWELBOs converge to their average, while $\Delta_L$ is more or less unchanged.}
    \label{fig:NLL_v_L_mnist}
\end{figure}

\begin{table*}[!t]
\centering
\caption{NLL results on six phylogenetic tree inference benchmark datasets for VBPI-NF with RealNVP, $S=5$, $L=1000$, $K=10$. The improvements obtained from using our framework (right column) are substantial in the field of phylogenetics (cf. Table 1 in \cite{zhang2018variational} or Table 1 in \cite{zhang2020improved}).}
\label{tab:vbpi_comparison}
\begin{tabular}{ c c c c c c} 
Dataset & Reference & Taxa & Sites & $\overline{\mathcal{L}}_L $& $\mathcal{L}_{\text{MIS}}^L$\\
\hline
DS1 & \citep{hedges1990tetrapod} & $27$ & $1949$ & $7108.42$ & $\mathbf{7108.10}$\\ 
DS2 & \citep{garey1996molecular} & $29$ & $2520$ & $ 26367.74$ & $\mathbf{26367.37}$\\ 
DS3 & \citep{yang2003comparison} & $36$ & $1812$ & $ 33735.15$ & $\mathbf{ 33734.89}$\\ 
DS4 & \citep{henk2003laboulbeniopsis} & $41$ & $1137$ & $13329.97$ & $\mathbf{13329.58}$\\ 
DS5 & \citep{lakner2008efficiency} & $50$ & $378$  & $8214.70$ & $\mathbf{8214.06}$ \\
DS8 & \citep{rossman2001molecular} & $64$ & $1008$ & $8650.73$ & $\mathbf{8650.32}$\\ 
\hline
\end{tabular}
\end{table*}

We now move to large-scale experiments, starting with the benchmark dataset MNIST  \citep{lecun1998mnist}. We experiment with deep ensembles of variational approximations using the state-of-the-art Nouveau VAE (NVAE;  \cite{vahdat2020nvae}) without NF. In order to obtain $f_{\phi_1}(x)$ and $f_\theta(z)$, we used the authors{'} exemplarily well-documented code, available on \url{https://github.com/NVlabs/NVAE}, training with the standard-ELBO-based objective as described in their work, by executing the commands provided there. Unfortunately we were not able to reproduce their results based on their descriptions\footnote{We tried different seeds and more training epochs. In the paper the authors reported an NLL score of $78.01$.}, however it is not critical for us to reproduce their results. Rather, we wish to compare MISELBO with IWELBO and ELBO for a state-of-the-art VAE. 

After training $\{f_{\phi_1}(x), f_\theta(z)\}$, we followed the algorithmic description in Algorithm \ref{alg:misvae} to get the ensemble of deep variational approximations. This means that we froze the NN weights in the decoder, $f_\theta(z)$, while initializing the NN weights in $\{f_{\phi_s}(x) : s\neq1\}$ randomly. Apart from these minor changes, the original code was not modified for our training, i.e. the architecture and training procedure was identical to what the authors had made available. This  illustrates the simplicity and versatility of our framework.

Next, we estimated the marginal log-likelihood using MISELBO (according to Algorithm \ref{alg:miselbo}), and reported the results in Table \ref{table:S=2 MNIST}, \ref{table:NLL-benchmarking} and in Figure \ref{fig:NLL_v_L_mnist}. Impressively, MISELBO consistently enhances the performance of the NVAE, i.e. $\Delta_L$ is strictly non-negative. In this experiment setting, we constrained ourselves to $S\in\{2, 3\}$, as our ablation study, performed on subsets of the data, showed that the JSD$(\mathcal{Q}_S)$ decreased for $S=4$ (see supplementary material). From Section \ref{sec:experiments:jsd} we have inferred that $\Delta_L$ might also decrease in this case, while $\overline{\mathcal{L}}_L$ does not appear to gain from increasing $S$.

In their paper,  \cite{vahdat2020nvae} use $L=1000$ importance samples to estimate the marginal log-likelihood. Indeed, in the MISELBO framework a total of $S\times L$ samples are drawn. However, when using MISELBO, far less importance samples are required in order to outperform NVAE using IWELBO. In Table \ref{table:S=2 MNIST} we demonstrate that NVAE with $L=50$ and $S=2$ still outperforms the IWELBO-based benchmarks we were able to obtain for NVAE with $L=1000$. The improvement in wall-clock time when comparing $\mathcal{L}_\text{MIS}^{50}$ ($S=2$) versus $\mathcal{L}_{1000}$ (note, not the average but a single NVAE) was remarkable. Averaged over four runs, $\mathcal{L}_\text{MIS}^{50}$ took $1605\pm3$ seconds, while $\mathcal{L}_{1000}$ needed $8054\pm3$ seconds. Hence, MISELBO used $90$\% less importance samples, was more than five times faster in computation time, and it was still superior to the IWELBO.

Ultimately, we highlight a novel insight: since non-zero JSD$(\mathcal{Q}_S)$ is obtained solely from random initialization of the NN weights in $\{f_{\phi_s}(x)\}$, this experiment also demonstrates that we can leverage deep ensemble diversity to translate the multimodal posterior distribution in the NN weight space \citep{wilson2020bayesian},  into a multimodal ensemble of variational approximations in the latent space.

\subsection{Phylogenetic Tree Inference}
\label{subsec:phylo_inf}
In contrast to the VAE considered above, the variational Bayesian phylogenetic inference (VBPI;  \cite{zhang2018variational}) framework does not train a generative model. Instead, $p_\theta(x|\lambda, \tau)$ is a  likelihood function commonly used in phylogeny, which can be  computed using the pruning algorithm presented in  \citet{felsenstein2004inferring}. We use the prior distributions over branch lengths
$\lambda$, and tree topologies, $\tau$, described in  \cite{zhang2020improved}; given the priors, the generative model has no free  parameters, i.e., there is no need to infer parameters  (see supplementary material for details).

We train an $S=5$ ensemble of variational approximations --- the tree topology encoders are subsplit Bayesian networks (SBNs;  \cite{zhang2018generalizing}) --- using $p_\theta(x,\lambda,\tau)$ for six real datasets, and experiment with MISELBO. The results are presented in table \ref{tab:vbpi_comparison} from evaluating the NLL scores of VBPI using NF when using MISELBO compared to the average of IWELBOs. The latter lower bound was the reported metric in the original paper, where they used $K=10$ flows of RealNVP \cite{dinh2016density}.

Utilizing the MISELBO framework consistently improves the marginal log-likelihood estimates. Note that $\overline{\mathcal{L}}_L$ and $\mathcal{L}_\text{MIS}^L$ use the same number of importance samples throughout the experiments. We stress that in the context of phylogenetics, these improvements are substantial (cf. Table 1 in \cite{zhang2018variational} or Table 1 in \cite{zhang2020improved}). The estimated JSDs in these experiments were all approximately $0.1$. Based on our reasoning in this work, this implies that the corresponding $\mathcal{Q}_S$ were not diverse (or at least far from the upper bound on JSD$(\mathcal{Q}_S)$, $\log5\approx 1.61$). This was not necessarily expected as the SBNs are not deep NNs. 

VBPI-NF employs a strongly non-Gaussian variational approximation in the tree topology space, $q_{\phi}(\tau|x)$, while the base-distribution over branch lengths for the NF, $q^0_\phi(\lambda|\tau, x)$, is a log-Normal distribution. The resulting $q_\phi^K(\lambda|\tau, x)$, i.e. after $K$ RealNVP flows, is highly expressive.
Therefore, this experiment does not only display that MISELBO improves density-estimation performances even when the variational approximations are non-Gaussian and/or expressive, it also shows that the framework is versatile enough to be applicable to modern, complex  VI methods. 

\section{Conclusion}
We have established the concept of deep ensembles of variational approximations and exciting connections between recent advances in IS and VI (MISELBO). These two contributions are the major components of our proposed framework, visualized in Figure \ref{fig:mappings}. We have shown that the framework is versatile, simple to apply for VI methods, and powerful, which we demonstrate by improving the density-estimation performances for two state-of-the-art models, NVAE and VBPI.

Moreover, we have provided a novel proof showing that MISELBO is tighter than the average of ELBOs. In practice, this result appears to generalize to the average of IWELBOs. Also, we are the first to measure the diversity of deep ensembles in the latent space of a latent variable model. This is a useful tool when, for instance, researching the role of the prior distribution in VAEs, a topic currently attracting much attention. The role of the prior distribution on the diversity of the ensemble is furthermore an interesting research path.

Finally, this work paves the way for new research in IS-based deep learning. In particular, we have incorporated recent advances in the IS literature, and we believe that many new connections and developments are ahead. For instance, adaptive IS (AIS) methods \citep{bugallo2017adaptive}, including gradient-based algorithms  \citep{elvira2015gradient,elvira2019langevin}, could be employed or developed ad-hoc for updating/refining the variational approximations and, thereby, also the ensemble weights. 

\section{Acknowledgements}
First, we acknowledge the insightful comments provided by the reviewers which have helped improve our work. This project was made possible through funding from the Swedish Foundation for Strategic Research grant BD15-0043, and from the Swedish Research Council grant 2018-05417\_VR. Some of the computations were enabled by resources provided by the Swedish National Infrastructure for Computing (SNIC), partially funded by the Swedish Research Council through grant agreement no. 2018-05973.

\bibliography{references.bib}


\clearpage
\appendix

\thispagestyle{empty}

\onecolumn \makesupplementtitle

Here we provide supplementary proofs and experimental details. We provide code for the experiments on GitHub: \url{https://github.com/Lagergren-Lab/MISELBO}. 

\section{Proofs}
Here we provide the proofs for Corollaries \ref{corollary strict inequality} and \ref{corollary equality}. Recall that, for both Corollaries, we consider $L=1$.

\subsection{Proof of Corollary \ref{corollary strict inequality}}
\label{appendix corollary strict inequality}
A condition for this Corollary to hold is that the supports of the distributions in $\mathcal{Q}_S$ are mutually disjoint. The assumption implies that, when $z\sim q_{\phi_s}(z|x)$
\begin{equation}
    \sum_{{s'}=1}^S q_{\phi_{s'}}(z|x) = q_{\phi_s}(z|x).
\end{equation}
We start by reformulating the JSD as follows
\begin{align}
\label{eq:jsd_supp}
    \text{JSD}(\mathcal{Q}_S) &= \mathbb{H}\left[
    \frac{1}{S}\sum_{s=1}^S q_{\phi_s}(z|x)
    \right] - \frac{1}{S}\sum_{s=1}^S\mathbb{H}\left[
    q_{\phi_s}(z|x)
    \right]\\
    &=  -\frac{1}{S}\sum_{s=1}^S\mathbb{E}_{q_{\phi_s(z|x)}}\left[
    \log {\frac{1}{S}\sum_{s=1}^S q_{\phi_s}(z|x)}
    \right] + \frac{1}{S}\sum_{s=1}^S\mathbb{E}_{q_{\phi_s(z|x)}}\left[
    \log {q_{\phi_s}(z|x)}
    \right]\\
    &= \log S + \frac{1}{S}\sum_{s=1}^S\mathbb{E}_{q_{\phi_s(z|x)}}\left[
    \log \frac{q_{\phi_s}(z|x)}{\sum_{{s'}=1}^S q_{\phi_{{s'}}}(z|x)}
    \right].
\end{align}
Using the assumption we, get that
\begin{equation}
    \frac{1}{S}\sum_{s=1}^S\mathbb{E}_{q_{\phi_s(z|x)}}\left[
    \log \frac{q_{\phi_s}(z|x)}{\sum_{{s'}=1}^S q_{\phi_{{s'}}}(z|x)}
    \right] = \frac{1}{S}\sum_{s=1}^S\mathbb{E}_{q_{\phi_s(z|x)}}\left[
    \log \frac{q_{\phi_s}(z|x)}{ q_{\phi_{{s}}}(z|x)}
    \right] = 0,
\end{equation}
and so we can complete the proof:
\begin{align}
\label{eq:diff=jsd}
    \Delta_1 = \text{JSD}(\mathcal{Q}_S) &= \log S + \frac{1}{S}\sum_{s=1}^S\mathbb{E}_{q_{\phi_s(z|x)}}\left[
    \log \frac{q_{\phi_s}(z|x)}{\sum_{{s'}=1}^S q_{\phi_{{s'}}}(z|x)}
    \right] \\
    &= \log S + \frac{1}{S}\sum_{s=1}^S\mathbb{E}_{q_{\phi_s(z|x)}}\left[
    \log \frac{q_{\phi_s}(z|x)}{q_{\phi_s}(z|x)}
    \right] = \log S > 0,
\end{align}
when $S > 1$.

\subsection{Proof of Corollary \ref{corollary equality}}
\label{appendix corollary equality}
For this Corollary, we instead assume that all variational approximations are identical, implying that
\begin{equation}
\label{eq:assumption 2}
    \sum_{s'=1}^S q_{\phi_{s'}}(z|x) = S q_{\phi_{s}}(z|x),
\end{equation}
when $z\sim q_{\phi_{s}}(z|x)$.

Using the reformulation of the JSD in Eq. \ref{eq:jsd_supp} and Eq. \ref{eq:assumption 2}, we have
\begin{align}
    \Delta_1 = \text{JSD}(\mathcal{Q}_S)
    &= \log S + \frac{1}{S}\sum_{s=1}^S\mathbb{E}_{q_{\phi_s(z|x)}}\left[
    \log \frac{q_{\phi_s}(z|x)}{\sum_{{s'}=1}^S q_{\phi_{{s'}}}(z|x)}
    \right]\\
    &= \log S + \frac{1}{S}\sum_{s=1}^S\mathbb{E}_{q_{\phi_s(z|x)}}\left[
    \log \frac{q_{\phi_s}(z|x)}{S q_{\phi_{{s}}}(z|x)}
    \right]\\ &= \log S + \frac{1}{S}\sum_{s=1}^S\mathbb{E}_{q_{\phi_s(z|x)}}\left[
    \log \frac{q_{\phi_s}(z|x)}{ q_{\phi_{{s}}}(z|x)}
     - \log S \right] = \log S - \log S = 0,
\end{align}
and so the equality in Corollary \ref{corollary equality} holds.

\section{Experiment \ref{subsec:ens_var_app} Details}
For the experiment using $p_1(z)$ as target distribution, we initialized $\mu_1$ to $(-3, 0)$ and $\mu_2$ to $(3,0)$. For the experiment using $p_2(z)$ as target distribution, we initialized $\mu_1$ to $(-3, 0)$, $\mu_2$ to $(0,0)$ and $\mu_3$ to $(3,0)$. The co-variance matrix of each variational distribution was fixed to $\sigma^2 I$ where $\sigma = 0.8$ and $I$ is the identity matrix of size 2. We trained each variational distribution for 10000 iterations, sampling 1000 $z$'s in each iteration. We used a learning rate of $0.001$ for the Adam optimizer. The training seed was set to 0.

We evaluated our models on 10000 samples using seed $=1$.

\section{Experiment \ref{sec:experiments:jsd} Details}

\subsection{Non-Hierarchical Case}
We consider three variants of the true distribution, $p(z)$.

Setting (i): we let $p(z)$ be a uniform mixture of six Gaussians with $\mu\in\{-5, 0, 5, 10, 15, 20\}$ and $\sigma=0.5$ (for all components). This setting is included in the main text and visualized in Figure 3.

\begin{figure}[H]
    \centering
    \includegraphics[width=1\textwidth]{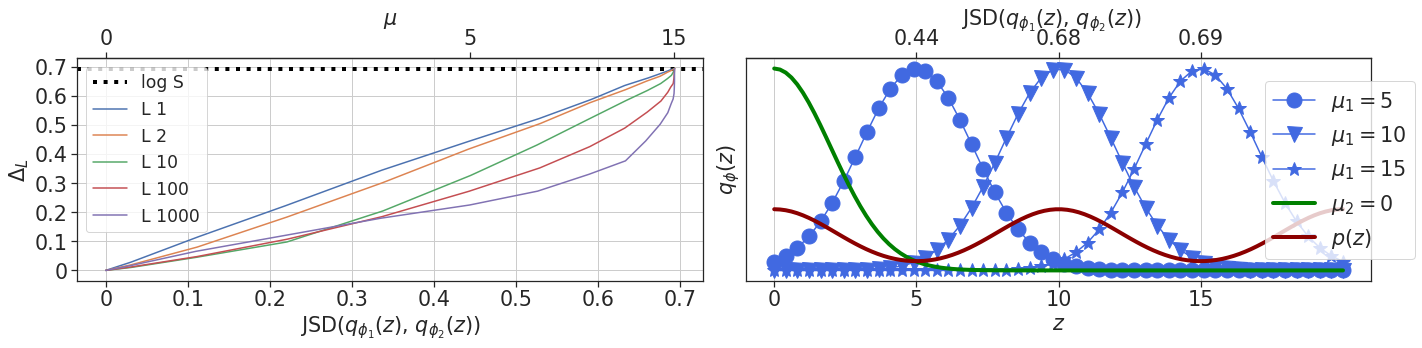}
    \caption{Setting (ii): trimodal Gaussian distribution and the corresponding JSD$(\mathcal{Q}_S)$.}
    \label{fig:trimodal}
\end{figure}

Setting (ii): we let $p(z)$ be a uniform mixture of three Gaussians with $\mu\in\{0, 10, 20\}$ and $\sigma=1.1$ (for all components). We visualize this setting here, in Figure \ref{fig:trimodal}.

\begin{figure}[H]
    \centering
    \includegraphics[width=1\textwidth]{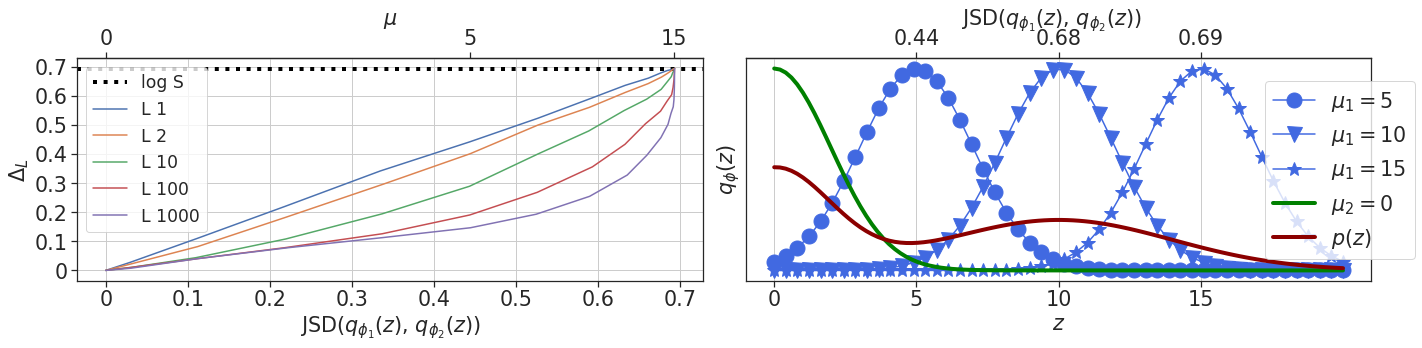}
    \caption{Setting (iii): bimodal Gaussian distribution and the corresponding JSD$(\mathcal{Q}_S)$.}
    \label{fig:bimodal}
\end{figure}

Setting (iii): we let $p(z)$ be a uniform mixture of two Gaussians,
\begin{equation}
p(z) =  \frac{1}{2}\mathcal{N}(z|0, 4) + \frac{1}{2} \mathcal{N}(z|10, 16).
\end{equation} 
We visualize this setting here, in Figure \ref{fig:bimodal}.

\subsection{Hierarchical Case}
We also considered a hierarchical model, which is visualized in the main text (see Figure \ref{fig:jsd_v_std}). We let
$
    p(z, \mu) = p(z)p(\mu),
$
where
$
    p(\mu) = \mathcal{N}(10, 9)
$
and $p(z)$ is the same as in setting (i) above.

\section{MNIST Experiment (\ref{sec:mnist}) Details}
When using the NVAE \citep{vahdat2020nvae} model, we first trained $f_{\phi_1}(x)$ and $f_\theta(z)$ with the same hyperparameters and \href{https://github.com/NVlabs/NVAE}{code} as in the original paper \citep{vahdat2020nvae} for five different seeds: $1, 2, 3, 4$ and $5$ ($1$ is used for the results in the original NVAE paper). The only exemption was related to the hardware: we used a single 32-GB Tesla V100 GPU, instead of two 16-GB Tesla V100 GPUs. 

A complete description of the hyperparameters can be found in Table 6 of Appendix A in \cite{vahdat2020nvae}, and in the MNIST experiment README at \url{https://github.com/NVlabs/NVAE}.

\textbf{Training seeds - } For the VAEs  $\{f_{\phi_1}(x), f_\theta(z)\}$ trained with seeds $2,3,4,5$, we used seed $=0$ when training each corresponding $f_{\phi_2}(x)$. For $\{f_{\phi_1}(x), f_\theta(z)\}$ trained with seed $1$, we used seed $=2$ for $f_{\phi_2}(x)$; 
Where appropriate, we trained $f_{\phi_3}(x)$ with seed $=3$.

\textbf{Evaluation seed -} We used seed $= 0$ during evaluation.

\subsection{Ablation Study}

In our ablation study, we used the same three models as in section 5.2. We then trained an additional encoder (using seed $=4$) following the same scheme as for $f_{\phi_2}(x)$ and $f_{\phi_3}(x)$, making a total of $S=4$. These were all trained using the entire MNIST training set. Due to computational restrictions, we then performed evaluations on eight subsets of the test data, $100$ samples a time. The samples in the subsets were indexed by $\{1000-1099, 1100-1199, ..., 1700-1799\}$. 

Finally, we calculated $\mathcal{L}^L_\text{MIS}$, JSD($\mathcal{Q}_S)$ and $\overline{\mathcal{L}}_L$ for each subset, reporting the mean and standard deviation over the subsets for each quantity. The results are presented in Table \ref{tab:ablation_study_table}. The entries are the means, and the standard deviations are in parentheses. We set seed $=0$ for all evaluations in the study.

Observing the values in the table, we note that the JSD$(\mathcal{Q}_S)$ decreases as we go from $S=3$ to $S=4$. As we found in general that $\Delta_L$ decreases with the JSD$(\mathcal{Q}_S)$, we decided, based on our ablation study, not to perform the full experiment with $S=4$ for computational reasons. Indeed, the largest $\Delta_L$ and best (mean) $\mathcal{L}^L_\text{MIS}$ were achieved when $S=3, L=50$ (bold entry in Table \ref{tab:ablation_study_table}). Additionally, note that the JSD$(\mathcal{Q}_S)$ is estimated using $L\times S$ importance samples, and averaged over the data samples. This is the explanation for varying JSDs when they should, in theory, be independent of $L$. 

\begin{table*}[!ht]
\centering
\caption{Mean and standard deviation of  $\mathcal{L}^L_\text{MIS}$, JSD($\mathcal{Q}_S)$ and $\overline{\mathcal{L}}_L$. Results from our ablation study on eight subsets of size 100 MNIST data using NVAE.}.
\begin{tabular}{ll|lll}
\label{tab:ablation_study_table}
   $S$ &     $L$ &  $\mathcal{L}^L_\text{MIS}$ &  JSD($\mathcal{Q}_S)$ & $\overline{\mathcal{L}}_L$ \\ \hline
2 &  1 &     79.46(2.055) & 0.22(0.021) &       79.67(2.059) \\
2 &  2 &     78.79(2.057) & 0.26(0.007) &       79.07(2.055) \\
2 &  3 &     78.55(2.086) & 0.22(0.012) &       78.82(2.079) \\
2 &  5 &     78.28(2.050) & 0.25(0.005) &       78.56(2.051) \\
2 & 10 &     78.02(2.065) & 0.25(0.006) &       78.30(2.065) \\
2 & 50 &     77.73(2.046) & 0.25(0.003) &       78.01(2.054) \\
3 &  1 &     79.42(2.040) & 0.29(0.028) &       79.70(2.045) \\
3 &  2 &     78.76(2.084) & 0.29(0.013) &       79.07(2.091) \\
3 &  3 &     78.50(2.065) & 0.28(0.011) &       78.81(2.058) \\
3 &  5 &     78.24(2.075) & 0.28(0.006) &       78.54(2.074) \\
3 & 10 &     78.00(2.045) & 0.29(0.005) &       78.31(2.050) \\
3 & 50 &     \textbf{77.71}(2.047) & 0.28(0.002) &       78.01(2.054) \\
4 &  1 &     79.41(2.066) & 0.26(0.020) &       79.67(2.065) \\
4 &  2 &     78.77(2.066) & 0.27(0.012) &       79.04(2.068) \\
4 &  3 &     78.53(2.058) & 0.27(0.009) &       78.82(2.054) \\
4 &  5 &     78.26(2.070) & 0.27(0.009) &       78.54(2.074) \\
4 & 10 &     78.03(2.042) & 0.27(0.005) &       78.31(2.045) \\
4 & 50 &     77.75(2.040) & 0.27(0.004) &       78.02(2.044) \\
\end{tabular}
\end{table*}

\newpage 
\section{VBPI-NF Experiment (\ref{subsec:phylo_inf}) Details}

\textbf{VBPI-NF Details:}

\begin{itemize}
    \item Version: Repository cloned at 7 May 2021. \url{https://github.com/zcrabbit/vbpi-nf}
    \item Flow Type (flow\_type): realnvp
    \item Number of layers for permutation invariant flow (lnf): 10
    \item Step size for branch length parameters (stepszBranch): 0.0001 (we consulted with the the author, 14 May 2021)
    \item Rest of the parameters are used with their default settings. 
    \item We modified the code so that we can fix the seed to reproduce the results.
\end{itemize}

Example script: 

\fbox{\begin{minipage}{0.9\textwidth}
python main.py --dataset data\_name --flow\_type realnvp --Lnf 10 --stepszBranch 0.0001 --vbpi\_seed 1
\end{minipage}}

Vbpi-Nf requires bootstrap trees to construct CPTs. The bootstrap trees for DS[1-4] are available in \url{https://github.com/zcrabbit/vbpi-nf}. For DS5, DS6 and DS8, we used UFBoot to create the bootstrap trees.

\begin{itemize}
    \item IQ-TREE Version: 1.6.12 \url{http://www.iqtree.org/}
    \item Number of independent runs: 10
    \item Model (m): JC69
    \item Number of bootstrap replicates (bb): 10,000
\end{itemize}

Example script: 

\fbox{\begin{minipage}{0.5\textwidth}
iqtree -s dataset\_name -bb 10000 -wbt -m JC69 -redo
\end{minipage}}

\newpage
\textbf{Lower bound details:}

\begin{itemize}
    \item For each dataset, VBPI-NF is run with 5 different seeds  ($S=5$) independently (used seeds values are $\{4,15,23,42,108\}$). 
    \item For each trained model $s$, we sampled $L=1000$ trees ($\tau_{1:L}$) and base branch lengths ($\lambda^{(0)}_{1:L}$). 
    \item We used the same tree and branch length samples to compute the IWELBOs and MISELBO. Note that, in normalizing flows, one \textit{samples} from the base distribution, whereas the final variables are obtained via a deterministic series of transformations. Here, this means that we sample base branch lengths, $\lambda^{(0)}_s$, from $q^{(0)}_{\phi_s}$. The final branch lengths, $\lambda^{(K+1)}_{s'}$, are not sampled, but obtained via the $s'$th model's normalizing flows.
\end{itemize}

Next we provide the expressions for the average IWELBOs and the MISELBO for this experiment. They are useful in order to understand how to apply normalizing flows in our framework.

\textbf{IWELBO for VBPI-NF}

\begin{equation}
   \overline{\mathcal{L}}_L = \frac{1}{S} \sum_{s=1}^S \mathbb{E}_{q_{\phi_s}(\tau_{1:L}, \lambda^{(0)}_{1:L} | x)}
    \left[\log\frac{1}{L}\sum_{\ell=1}^L 
    \frac{p_\theta\big(x|\tau_{s,\ell},\lambda^{(K+1)}_{s,\ell}\big)
    p_\theta\big(\tau_{s,\ell},\lambda^{(K+1)}_{s,\ell}\big) }
    { q_{\phi_s}(\tau_{s,\ell}) 
    q^{(0)}_{\phi_s}\big(\lambda^{(0)}_{s,\ell} | \tau_{s,\ell} \big) 
    \prod_{k=0}^K \big| \textnormal{det} \frac{\partial(\lambda^{(k+1)}_{s,\ell})}{\partial(\lambda^{(k)}_{s,\ell})} \big|^{-1}  }
    \right].
\end{equation}

\textbf{MISELBO for VBPI-NF}

\begin{equation}
    \mathcal{L}_\text{MIS}^L = \frac{1}{S} \sum_{s=1}^S \mathbb{E}_{q_{\phi_s}(\tau_{1:L}, \lambda^{(0)}_{1:L} | x)}
    \left[\log\frac{1}{L}\sum_{\ell=1}^L 
    \frac{p_\theta\big(x|\tau_{s,\ell},\lambda^{(K+1)}_{s,\ell}\big)
    p_\theta\big(\tau_{s,\ell},\lambda^{(K+1)}_{s,\ell}\big) }
    {\frac{1}{S}\sum_{s'=1}^S q_{\phi_{s'}}(\tau_{s,\ell}) 
    q^{(0)}_{\phi_{s'}}\big(\lambda^{(0)}_{s,\ell} | \tau_{s,\ell} \big) 
    \prod_{k=0}^K \big| \textnormal{det} \frac{\partial(\lambda^{(k+1)}_{s',\ell})}{\partial(\lambda^{(k)}_{s',\ell})} \big|^{-1}  }
    \right].
\end{equation}

\textbf{Comment on the generative model:} In the associated section in the main text, we state that there are no free (read \textit{learnable}) parameters in the generative model, $p_{\theta}(z, \lambda, \tau)$. Meanwhile, we parameterize $p$ by $\theta$ in order to emphasize that there are indeed model assumptions: The likelihood function assumes an evolutionary substitution model (JC69). The prior on branch lengths, $\lambda$, is assumed to be an exponential distribution, $p_{\theta}(\lambda)=\text{Exp}(10)$. The prior on the tree topologies, $\tau$, is assumed to be a uniform distribution over the space of unrooted binary trees, $p_{\theta}(\tau) = \left(\frac{(2n - 5)!}{2^{n - 3} (n - 3)! }\right)^{-1}$, where $n\geq 3$ are the number of taxa. When $n<3$ there exists only a single topology.

All the above assumptions are the same as in \cite{zhang2020improved}.

\end{document}